\documentclass[sigconf]{acmart}
\AtBeginDocument{%
  }

\copyrightyear{2026}
\acmYear{2026}
\setcopyright{cc}
\setcctype{by-nc-nd}
\acmConference[DAC '26]{63rd ACM/IEEE Design Automation Conference}{July 26--29, 2026}{Long Beach, CA, USA}
\acmBooktitle{63rd ACM/IEEE Design Automation Conference (DAC '26), July 26--29, 2026, Long Beach, CA, USA}
\acmDOI{10.1145/3770743.3804292}
\acmISBN{979-8-4007-2254-7/2026/07}

\acmSubmissionID{2199}

\usepackage[ruled,vlined]{algorithm2e}
\usepackage{xcolor}
\usepackage{amsmath}
\usepackage{enumitem}
\usepackage{multirow,multicol}
\usepackage{subfigure}
\usepackage{makecell}
\usepackage{pifont}
\newcommand{\name}{\textit{ExpertFlow}} 
\begin{document}

\title{{\name}: Efficient Mixture-of-Experts Inference via Predictive Expert Caching and Token Scheduling}


\author{Xin He$^{1_\textsuperscript{*}}$,
Shunkang Zhang$^{2}$,
Kaijie Tang$^{3}$,
Shaohuai Shi$^{3}$,
Yuxin Wang$^{4}$,
Zihao Zeng$^{5}$,\\
Zhenheng Tang$^{2}$,
Xiaowen Chu$^{6}$,
Haiyan Yin$^{1}$,
Ivor W. Tsang$^{1,5}$,
Yew Soon Ong$^{1,5}$}\authornote{Corresponding authors: \{he\_xin,ong\_yew\_soon\}@a-star.edu.sg}

\affiliation{%
$^{1}$CFAR, Agency for Science, Technology and Research (A*STAR), Singapore \\
$^{2}$The Hong Kong University of Science and Technology, Hong Kong \\
$^{3}$Harbin Institute of Technology, Shenzhen, China  \quad
$^{4}$Hong Kong Baptist University, Hong Kong  \\
$^{5}$Nanyang Technological University, Singapore \\
$^{6}$The Hong Kong University of Science and Technology (Guangzhou), China 
\country{}
}

\renewcommand{\shortauthors}{He et al.}

\renewcommand{\authors}{Xin He$^{1_\textsuperscript{*}}$,
Shunkang Zhang$^{2}$,
Kaijie Tang$^{3}$,
Shaohuai Shi$^{3}$,
Yuxin Wang$^{4}$,
Zihao Zeng$^{5}$,
Zhenheng Tang$^{2}$,
Xiaowen Chu$^{6}$,
Haiyan Yin$^{1}$,
Ivor W. Tsang$^{1,5}$,
Yew Soon Ong$^{1,5_\textsuperscript{*}}$}

\begin{abstract}

Sparse Mixture-of-Experts (MoE) models can outperform dense large language models at similar computation by activating only a small set of experts per token. However, stacking many expert modules introduces substantial parameter memory, which makes MoE models difficult to deploy in memory-constrained environments such as single-GPU devices. Offloading alleviates this issue by storing inactive experts in CPU memory and loading them on demand, but existing methods remain limited: static caches disregard input-dependent routing, and methods that train separate models to predict expert usage ahead of time are often inaccurate or require significant training cost. We propose \textbf{ExpertFlow}, a lightweight MoE inference system that addresses this routing dependency through three coordinated components: 1) a transformer-based routing path predictor that estimates expert usage across all MoE layers in a single forward pass, 2) a token scheduler that groups tokens with similar predicted routes to improve expert utilization, and 3) a predictive expert cache that loads only the required experts while correcting mispredictions at runtime. Together, these components enable efficient expert loading and execution, reducing GPU memory usage by up to 93.72\% and improving inference throughput by up to 10$\times$ over strong offloading baselines on a single GPU.

\end{abstract}




\maketitle

\section{Introduction}
Sparse MoE models~\cite{shazeer2017outrageouslyMoE, fedus2022switchtransformer, dai2024deepseekmoe, jiang2024mixtral} scale parameter size efficiently by activating only a small subset of experts per input, reducing per-token computation while keeping accuracy comparable to dense LLMs~\cite{gpt2, t5, guo2024deepseek}. This efficiency, however, increases memory usage. For instance, Mixtral-8$\times$7B~\cite{jiang2024mixtral} requires more than 96 GB of GPU memory, exceeding the 80 GB capacity of an NVIDIA A100 GPU. Although 45.1B of its 46.7B parameters belong to expert modules, only a small fraction is used per input, leading to substantial memory redundancy. This sparsity suggests that offloading inactive experts to CPU and loading only the needed ones can reduce GPU memory demand. Existing studies have investigated such offloading methods~\cite{eliseev2023moe-offload, kamahori2024fiddler, hwang2023pregatemoe, shen2022se-moe, du2024sida, song2024promoe}, but remain limited by three challenges.


\textbf{Inefficient expert prediction.} Early and accurate expert activation prediction underpins effective offloading, as it enables scheduling and prefetching before experts are required. Prior work takes two routes. \textit{Regression-based methods}~\cite{hwang2023pregatemoe,du2024sida} approximate router scores, but even small score errors can affect output quality, necessitating extensive fine-tuning to recover the original routing. \textit{Classification-based methods} predict selected experts directly. Heuristic variants based on token–expert statistics~\cite{xue2024openmoe,jiang2024mixtral,lina} are lightweight but fail to capture input-dependent routing behavior. Learning-based predictors (e.g., ProMoE~\cite{song2024promoe}) improve accuracy, yet their layer-by-layer sequential design reveals expert usage only after the previous layer executes, restricting scheduling flexibility.


\begin{figure}[!tb]
    \centering
    \includegraphics[width=\linewidth]{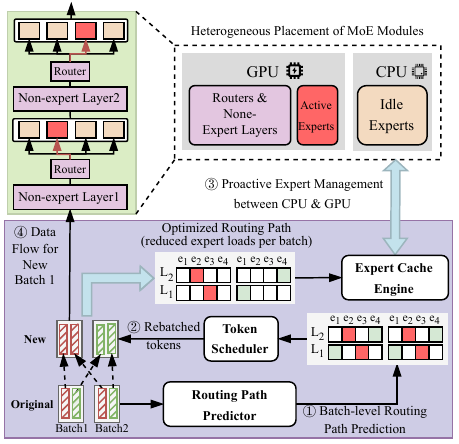}
    \caption{Overview of {\name}. Given two input batches, \ding{192}  \textbf{Routing Path Predictor} predicts expert activation across all layers for all tokens, 
\ding{193}  \textbf{Token Scheduler} uses the prediction to reorder tokens across batches to consolidate expert usage, 
\ding{194}  \textbf{Expert Cache Engine} preloads only required experts into GPU from CPU, and 
\ding{195} the MoE model executes with optimized token flow and heterogeneous expert placement. 
}
    
    \label{fig:overview}
\end{figure}
\textbf{Low expert utilization.} In the decoding phase, the token distribution across experts can be highly imbalanced, and some experts may receive only a single token. Since expert kernels have near-constant cost when handling a small number of tokens~\cite{williams2009roofline}, such sparse assignments lead to low compute efficiency.


\textbf{Ineffective expert caching.} Expert caching is central to controlling GPU memory usage. The commonly used LRU policy~\cite{eliseev2023moe-offload} evicts experts purely by recency and overlooks routing patterns, leading to unstable cache hit rates under MoE’s dynamic activations. SE-MoE~\cite{shen2022se-moe} improves locality by caching all experts from two consecutive layers through a ring-buffer design, but this creates large memory overhead for models with many experts (e.g., Switch-128) and repeatedly loads inactive experts, resulting in unnecessary CPU–GPU transfers.


To address these challenges under resource-constrained settings such as single-GPU inference, we propose {\name}, a unified system for memory-efficient MoE execution. Fig.~\ref{fig:overview} shows an example where tokens from two batches activate different experts across layers, leading to fragmented execution and high memory usage when processed directly. {\name} recasts this process as a predictive and coordinated pipeline through three components: \ding{172} the \textbf{Routing Path Predictor (RPP)} predicts expert activations for all tokens and all layers in one forward pass, providing early global routing signals; \ding{173} the \textbf{Token Scheduler (TS)} reorganizes tokens based on predicted paths to consolidate expert usage and increase compute efficiency; \ding{174} the \textbf{Expert Cache Engine (ECE)} loads only the needed experts into GPU memory and reuses them across steps, with lightweight correction for mispredictions. Our contributions are summarized as follows:


\begin{itemize}[leftmargin=*]
\item We identify three core bottlenecks in MoE offloading: inefficient expert prediction, low expert utilization, and ineffective caching under dynamic routing.

\item We introduce {\name}, a unified system that integrates predictive scheduling, routing-aware token rebatching, and adaptive caching with lightweight correction. {\name} reduces GPU memory usage by up to 93.72\% and improves throughput by up to 10$\times$ over strong offloading methods, enabling efficient MoE inference on constrained single-GPU settings.

\item Our \textit{RPP} achieves up to 95\% expert prediction accuracy with strong cross-domain generalization. The \textit{TS} improves throughput by up to 16.19\% via enhanced expert reuse. The \textit{ECE} attains a cache hit ratio of 91.96\%, outperforming LRU by up to 61.15\%.

\end{itemize}

\section{Related Work}

\subsection{Mixture-of-Experts (MoE)}

MoE models~\cite{jacobs1991adaptivemoe} improve scalability by activating only a subset of experts for each token through a softmax-based gating mechanism, where the gating score for experts is $G(x)=\text{softmax}(xW_g)$ and the model selects the top-$k$ experts with the highest scores. The MoE layer output is then computed as a weighted sum of the selected experts, $y=\sum_{i\in\text{TopK}(G(x))} G_i(x)\,E_i(x)$. With advances in hardware and training methods, transformer-based MoE architectures have become widely used and show strong performance across many tasks~\cite{fedus2022switchtransformer,dai2024deepseekmoe,shazeer2017outrageouslyMoE,tang2024fusefl}, where the gating function determines which experts each token activates and shapes the routing pattern that drives system efficiency.



\subsection{Model Compression} 

LLM inference faces substantial GPU memory constraints, prompting prior research to explore a range of solutions. Distillation techniques~\cite{shen2022se-moe,dai2024deepseekmoe} reduce the number of experts by compressing the teacher network into a smaller student network. Model pruning methods have also been explored, such as pruning non-essential experts during fine-tuning based on usage frequency~\cite{chen2022taskmoe} and merging similar experts followed by low-rank decomposition~\cite{mc-smoe}. Post-training quantization~\cite{eliseev2023moe-offload,li2024quanBenchmark,lin2024awq,frantar2022gptq} further reduces memory consumption by converting pre-trained models to lower-precision ones (e.g., Int4), without requiring extensive retraining. The contribution of our proposed {\name} is orthogonal to this direction, and {\name} can be seamlessly integrated with these techniques to further reduce GPU memory cost during MoE inference.

\subsection{Model Offloading}

Model offloading reduces GPU memory usage by moving model states or computations to cheaper storage or processing units. Early work such as ZeRO~\cite{rajbhandari2020zero,ren2021zero-offload} offloaded optimizer states, gradients, and weights to CPUs or SSDs during training, and later extensions applied similar ideas to inference~\cite{song2023powerinfer,sheng2023flexgen,chen2024attentionOffload}. FlexGen~\cite{sheng2023flexgen} uses a zig-zag block schedule to offload activations and KV caches, allowing large models like OPT-175B~\cite{zhang2022opt} to run on a single 16GB GPU, while Lamina~\cite{chen2024attentionOffload} improves efficiency by shifting attention computation to CPUs. However, these methods are designed for dense LLMs and do not handle the dynamic, input-dependent routing of MoE models. Existing MoE offloading approaches either rely on low-accuracy heuristics~\cite{lina,eliseev2023moe-offload} or require costly predictor training~\cite{hwang2023pregatemoe,du2024sida}, limiting practical adoption. In contrast, we develop a unified system that provides accurate and low-cost expert routing prediction, enabling more efficient and flexible MoE inference.

\section{Method}

\subsection{System Design Overview}

Deploying MoE models under tight memory budgets, such as single-GPU settings, is challenging due to dynamic expert routing, low expert utilization, and heavy parameter movement. These issues are interdependent: inaccurate routing prediction limits scheduling, fragmented token batches activate unnecessary experts, and poor memory planning amplifies transfer overhead. {\name} addresses these challenges through a unified design that integrates a routing path predictor (\textit{RPP}), a token scheduler (\textit{TS}), and an expert cache engine (\textit{ECE}). The \textit{RPP} provides early global routing signals, enabling the \textit{TS} to reorganize tokens by predicted paths and the \textit{ECE} to prefetch only the required experts. This prediction-informed coordination jointly optimizes expert execution, data movement, and memory usage, enabling efficient and scalable MoE inference on constrained hardware.

\begin{figure}[!htb]
    \centering    
    \includegraphics[width=\linewidth]{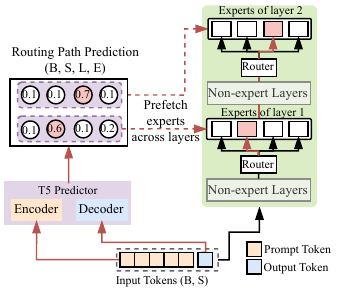}   
    \caption{Routing Path Predictor (RPP). Given a batch of $B$ sequences, each with $S$ input tokens, RPP predicts MoE expert activations across $L$ layers and $E$ experts in a single pass. It outputs activation probabilities of shape $(B,S,L,E)$ to support early expert prefetching.}
    \label{fig:predictor}
\end{figure}

\subsection{Routing Path Predictor (RPP)}\label{sec:moe_predictor}

\subsubsection{Predictor Architecture}

Existing predictors~\cite{song2024promoe,hwang2023pregatemoe} use MLPs to infer expert choices layer by layer, creating a sequential dependency that prevents early scheduling and prefetching. {\name} replaces this design with a T5-style encoder--decoder architecture~\cite{t5} (Fig.~\ref{fig:predictor}). The encoder embeds the full input sequence, and the decoder generates routing predictions in one pass. We attach $L$ lightweight heads to the decoder, each producing logits over $E$ experts for a specific MoE layer. This architecture exposes the complete routing plan before the first MoE layer executes, enabling early prefetching and coordinated memory planning.

\subsubsection{Predictor Training}

The predictor is trained to produce accurate routing paths for all $L$ MoE layers, each comprising $E$ experts, in a single forward pass. We log each token’s expert selections and encode its routing path as a binary matrix $r \in \{0,1\}^{L \times E}$. The predictor outputs a probability matrix $p$ of the same shape, and training is formulated as a multi-label classification task using binary cross-entropy:
\begin{equation}
\mathcal{L}=\frac{1}{LE}\sum_{l=1}^{L}\sum_{e=1}^{E}\left[r_{l,e}\log p_{l,e}+(1-r_{l,e})\log(1-p_{l,e})\right].
\end{equation}

\subsection{Token Scheduler (TS)}
\label{sec:token_scheduler}

\begin{figure}[!hbt]
    \centering
    \includegraphics[width=\linewidth]{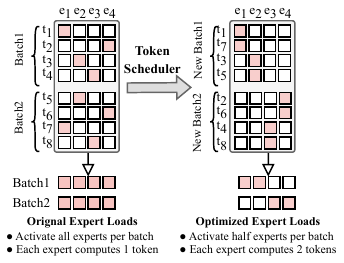}
    \caption{Token Scheduler (TS). Left: normal batch inference routes tokens to different experts, producing a worst-case pattern where all experts are active with only one token. Right: TS groups tokens with similar routing path into new batches, reducing active experts and increasing per-expert token load for better efficiency.}
    \label{fig:token_scheduler}
\end{figure}

In the decoding stage, an adverse routing pattern can arise where each small batch activates almost all experts while each expert receives only one token. Fig.~\ref{fig:token_scheduler} (left) illustrates this worst case for a single MoE layer with four experts and two batches of size four: tokens in each batch select different experts, so every batch activates all experts, causing frequent expert swapping and low per-expert workload. To address this, we introduce the \textit{Token Scheduler (TS)}, which rebatches tokens between two consecutive batches and groups tokens with similar expert selections into the same batch, as shown in Fig.~\ref{fig:token_scheduler} (right). This rebatching reduces the number of active experts per batch while increasing the tokens per expert, improving cache reuse and GPU efficiency.

\subsubsection{Mathematical Formulation}
\label{subsubsec:math_formulation}

Each token’s routing path is encoded as a binary matrix $r_i \in \{0,1\}^{L \times E}$. For two adjacent batches with $T$ tokens each, we merge all $2T$ tokens into a global set $\mathcal{T} = \{1,2,\ldots,2T\}$ and seek to split it into two disjoint batches $\mathcal{T}_1$ and $\mathcal{T}_2$ of equal size. The routing matrices for the new batches, $R_1$ and $R_2$, are obtained by applying an element-wise logical OR $\bigvee$ over the routing paths of the tokens assigned to each batch $R_1 = \bigvee_{i \in \mathcal{T}_1} r_i,\, R_2 = \bigvee_{i \in \mathcal{T}_2} r_i$. The objective is to minimize the total number of activated experts across both batches:

\begin{equation}
\min_{\mathcal{T}_1,\mathcal{T}_2}\;\sum_{l=1}^{L}\sum_{e=1}^{E}\big(R_1^{l,e}+R_2^{l,e}\big),
\label{eq:token_scheduler_obj}
\end{equation}
\noindent where $R_k^{l,e}=1$ indicating that expert $e$ at layer $l$ is activated in batch $k$. This objective explicitly promotes expert-wise co-location of similar tokens to reduce cache misses and raise per-expert load.

\subsubsection{K-Means Clustering for Fast Token Rebatching}
\label{subsubsec:kmeans_solution}

Solving Eq.~\ref{eq:token_scheduler_obj} exactly online is intractable, so we approximate it using a K-means-style clustering over routing-path similarity. For the $2T$ tokens, we construct a similarity matrix $S\in\mathbb{R}^{2T\times 2T}$, where $S_{ij}=1-\frac{d_{ij}}{L E}$ measures how close the routing paths of tokens $i$ and $j$ are, with $d_{ij}$ being their Hamming distance. We cluster tokens into two equal-size groups by iteratively assigning each token to the closest cluster under $S$ and updating centroids as the tokens with the highest average intra-cluster similarity. The procedure converges quickly or stops after a preset iteration limit, yielding $(\mathcal{T}_1,\mathcal{T}_2)$ as a fast approximation to Eq.~\ref{eq:token_scheduler_obj} with negligible CPU overhead (<10ms).

\subsubsection{Adaptive KV-Cache Management}

Rebatching perturbs token orderings assumed by the transformer’s key–value (KV) cache. \textit{TS} therefore incorporates two lightweight primitives to preserve attention semantics: \textbf{Merge}, which reconstructs the KV cache by aggregating token states according to global token order across the original batches; and \textbf{Reindex}, which updates token indices to the new layout for consistent KV lookup post-reordering. 

\subsubsection{Dual-Batch Inference Pipeline}\label{sec:dual-batch_inference}

\begin{figure}[!ht]
    \centering
    \includegraphics[width=\linewidth]{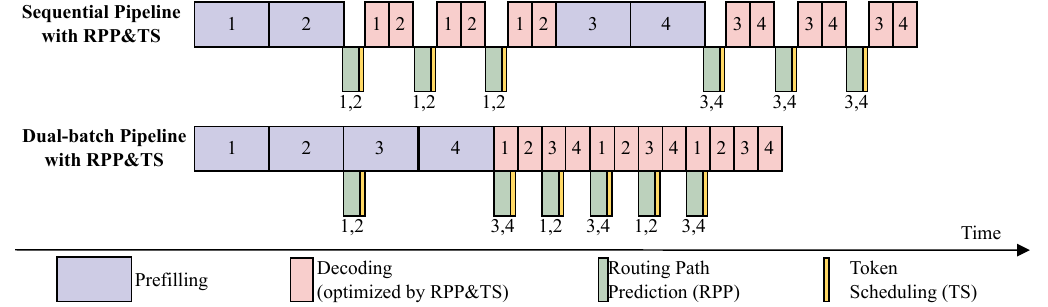}
    \caption{Sequential pipeline versus our Dual-Batch pipeline.}
    \label{fig:scheduler_overlap}
\end{figure}

We propose a \textit{Dual-Batch Inference Pipeline} (Fig.~\ref{fig:scheduler_overlap}) to hide the overhead of RPP and TS. The pipeline groups every two batches into one scheduling unit, which matches the requirement of TS to reorganize tokens across batches. During execution, model prefill and decoding for the current unit run in parallel with RPP and TS for the next unit. This interleaving avoids blocking and keeps the GPU well utilized.

\subsection{Expert Cache Engine (ECE)}\label{sec:ece}

The Expert Cache Engine (\textit{ECE}) manages expert parameters between GPU and CPU by combining two components: \textit{Predictive Locality-aware Expert Caching} (PLEC), which plans cache layout and prefetching based on predicted routing, and a \textit{Real-time Correction} mechanism that resolves prediction errors during execution. Together, they enable expert-level caching that is both proactive and adaptive to unexpected routing behaviors.

\subsubsection{Predictive Locality-aware Expert Caching (PLEC)}\label{sec:PLEC}

Unlike conventional cache policies such as LRU~\cite{eliseev2023moe-offload,yuan2024efficientLRU,aminabadi2022deepspeed-inference} and LFU~\cite{xu2024cachedLFU}, which operate without prediction and therefore use fixed per-layer cache allocations, \textit{PLEC} leverages routing predictions to adaptively assign cache slots across layers and prefetch the experts required in the next computation stage. This adaptive slot planning allows the cache to better match the anticipated expert demand, reducing unnecessary swaps and improving the effectiveness of prefetching.

\begin{figure}[!htb]
    \centering
    \includegraphics[width=\linewidth]{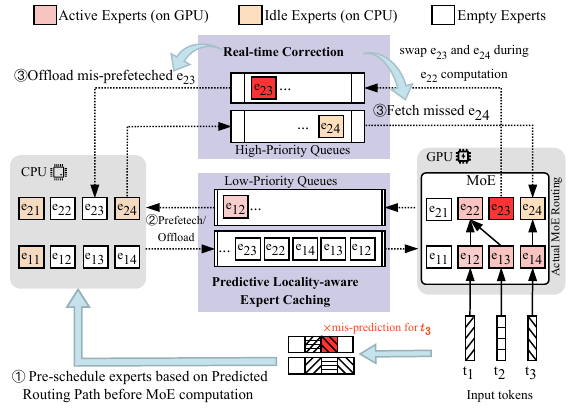}
    \caption{The workflow of \textit{Expert Cache Engine (\textit{ECE})}. \textit{ECE} pre-schedules experts to GPU based on routing predictions. During execution, it detects mispredictions (e.g., unwanted $e_{23}$ or missed $e_{24}$) and performs prioritized swaps while $e_{22}$ runs, overlapping I/O with compute to maintain throughput.}
    \label{fig:ece}
\end{figure}

As shown in Fig.~\ref{fig:ece}, consider an MoE model with two layers and four experts per layer. The predictor forecasts that tokens ${t_1,t_2,t_3}$ will activate three experts in layer-1 ($e_{12},e_{13},e_{14}$) and two in layer-2 ($e_{22},e_{23}$), while the GPU cache can hold only four experts. Since the predicted demand (five experts) exceeds the cache budget, PLEC allocates slots according to predicted usage—three to layer-1 and one to layer-2 (step \ding{172})—and prefetches the four most probable experts ($e_{12},e_{13},e_{14},e_{22}$) before execution (step \ding{173}). During computation, early-layer experts finish first and free their slots; these slots are then reused to load remaining predicted experts, such as loading $e_{23}$ once $e_{12}$ completes. This predictive allocation combined with runtime reuse reduces transfers and improves cache efficiency.

\section{Evaluation}\label{sec:perf_eval}

\subsection{Setup}\label{sec:exp_setup}

\textbf{Hardware Settings.} Our experiments were conducted on a single NVIDIA A40 GPU with 48 GB of memory and Intel(R) Xeon(R) Gold 6338 CPU @ 2.00GHz.

\textbf{Tasks and MoEs.} We evaluate on four datasets: Alpaca~\cite{alpaca} for chat, WMT16~\cite{wmt16} for translation, XSUM~\cite{xsum} for summarization, and AIME2024~\cite{aime2024} for problem solving. Our experiments cover six MoE models: Qwen1.5-MoE~\cite{qwen_moe}, Deepseek-MoE~\cite{dai2024deepseekmoe}, Mixtral-8$\times$7B~\cite{jiang2024mixtral}, and Switch Transformer with 32, 64, and 128 experts~\cite{fedus2022switchtransformer}. Model specifications are provided in Table~\ref{tab:moeModels}.



\textbf{Predictor Settings.} To evaluate robustness under domain shift, we consider two types of routing path predictors (\textit{RPP}). An \textit{in-domain} predictor is trained on the same dataset used for inference, while a \textit{cross-domain} predictor is trained on a different dataset. For example, when evaluating on WMT16, the in-domain \textit{RPP} is trained on WMT16, whereas the cross-domain \textit{RPP} is trained on XSUM.

\begin{table}
    \centering
    \scalebox{0.95}{
    \begin{tabular}{lcccc}
        \hline
         MoE  &\textbf{L} & \textbf{Act.P}/\textbf{P}  & \textbf{Act.E}/\textbf{E} & \textbf{E.P}  \\\hline
        Switch-32 & 12 & 0.22/1.98 B & 1/32 & 91.58\% \\
Switch-64 & 12 & 0.22/3.79 B & 1/64 & 95.61\% \\
Switch-128 & 12 & 0.22/7.41 B & 1/128 & 97.75\% \\
Mixtral-8 & 32 & 12.90/46.70 B & 2/8 & 96.57\% \\
Qwen1.5 & 24 & 2.70/14.30 B & 4/60 & 88.95\% \\
Deepseek-MoE & 27 & 2.80/16.40 B & 6/64 & 94.14\%\\
\hline
    \end{tabular}
    }
    \caption{MoE Configurations. \textbf{L}, \textbf{P}, and \textbf{E} denote layers, total parameters, and experts per layer. \textbf{Act.P} and \textbf{Act.E} refer to activated parameters and experts per token. \textbf{E.P} is the expert-to-total parameter ratio.}
    \label{tab:moeModels}
\end{table}

\begin{figure*}[!htb]
    \centering
    \subfigure[Switch Series (WMT16)]{
    \includegraphics[height=2.0cm]{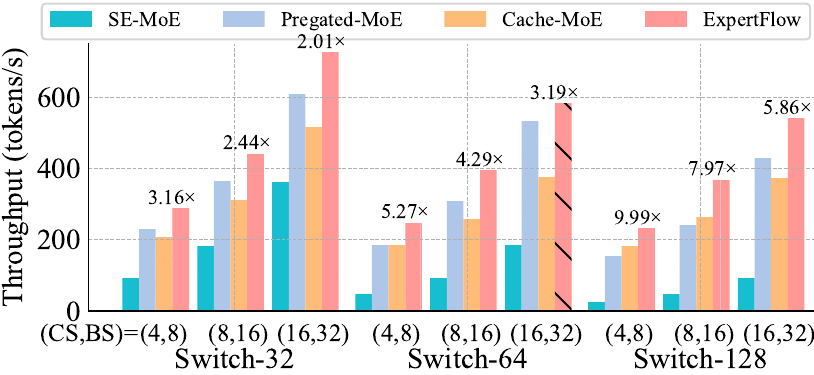}
    }
    \subfigure[Mixtral-8 (XSUM)]{    \includegraphics[height=2.0cm]{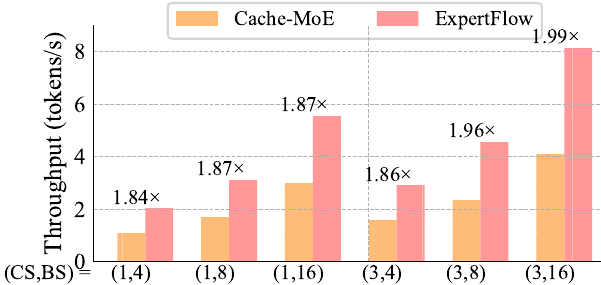}
    } 
    \subfigure[Qwen1.5 (Alpaca)]{
    \includegraphics[height=2.0cm]{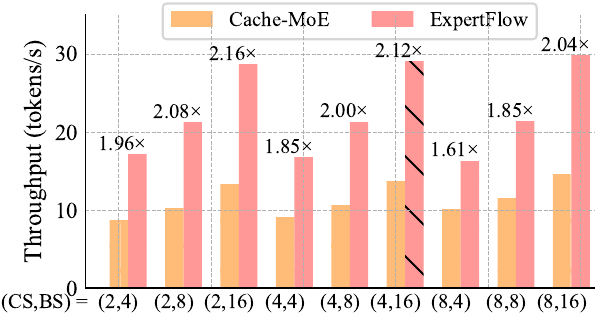}
    }    
    \subfigure[Deepseek-MoE (AIME2024)]{
    \includegraphics[height=2.0cm]{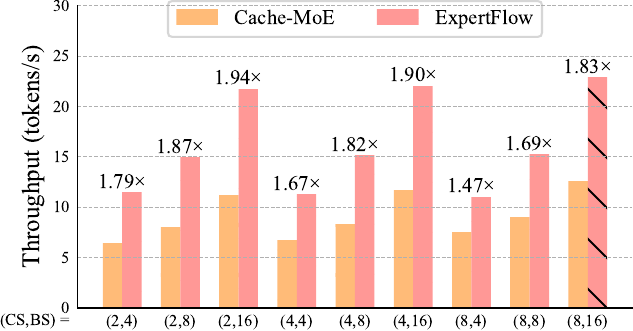}
    }
    \caption{Throughput across different MoE models and datasets.
Our results are obtained using \textbf{in-domain} predictors.}
    \label{fig:indomain_speedup}
\end{figure*}

\begin{figure}[!htb]
    \centering
    \subfigure[WMT16]{
    \includegraphics[width=.47\linewidth]{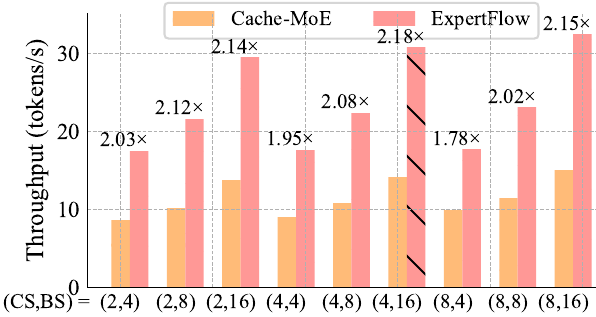}
    }
    \subfigure[XSUM]{
    \includegraphics[width=.47\linewidth]{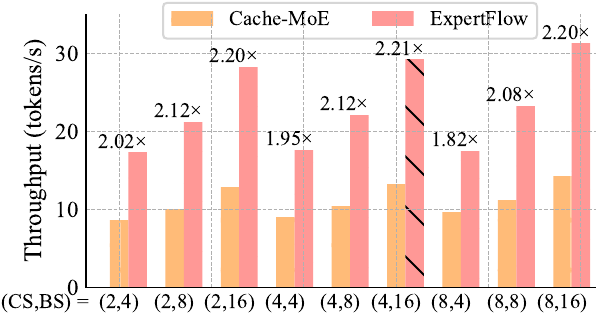}
    }    
    \caption{Throughput comparison for Qwen1.5 on WMT16 and XSUM datasets. Our results are obtained using a \textbf{cross-domain} predictor trained on Alpaca.}
    \label{fig:crossdomain_speedup}
\end{figure}

\subsection{Inference Performance}
\label{sec:speed_results}

\subsubsection{Baselines}
We compare against three representative methods. \textit{Cache-MoE}~\cite{eliseev2023moe-offload} maintains a fixed per-layer expert cache with \textbf{LRU} replacement, falling back to CPU on misses. \textit{SE-MoE}~\cite{shen2022se-moe} preloads experts for multiple layers and employs ring scheduling to overlap compute and data movement. \textit{Pregated-MoE}~\cite{hwang2023pregatemoe} trains MLP-based routers to select experts without runtime gating. All three are evaluated on Switch Transformers, which their implementations fully support. For Mixtral-8 and Qwen1.5, we report against \textit{Cache-MoE} only, as other baselines lack architectural support and router weights; under these conditions, \textit{Cache-MoE} is a strong baseline.

\subsubsection{In-domain Throughput}
We evaluate four model–dataset pairs (Switch on WMT16, Mixtral-8 on XSUM, Qwen1.5 on Alpaca and Deepseek-MoE on AIME2024) under varying cache size (\textbf{CS}) and batch size (\textbf{BS}) (Fig.~\ref{fig:indomain_speedup}). For the Switch series (Fig.~\ref{fig:indomain_speedup}a), gains increase with the number of experts: at (CS=16, BS=32), our method yields 2.01$\times$, 3.19$\times$, and 5.86$\times$ speedups over \textit{SE-MoE} on Switch-32/64/128, respectively. On Switch-128, tightening memory further amplifies benefits: reducing CS from 16 to 4 increases speedup from 5.86$\times$ to 9.99$\times$. For Mixtral-8, Qwen1.5 and Deepseek-MoE (Figs.~\ref{fig:indomain_speedup}b–d), throughput improves with larger BS due to \textit{TS}-enabled expert reuse; we outperform \textit{Cache-MoE} by up to 1.99$\times$ (Mixtral-8), 2.12$\times$ (Qwen1.5) and 1.94$\times$ (Deepseek-MoE) via accurate prefetching and reduced load latency.

\subsubsection{Cross-domain Throughput}
To assess robustness, we apply an \textit{RPP} trained on Alpaca to Qwen1.5 inference on XSUM and WMT16 (Fig.~\ref{fig:crossdomain_speedup}). Our method consistently surpasses \textit{Cache-MoE}, achieving up to 2.18$\times$ (WMT16) and 2.21$\times$ (XSUM) at (CS=4, BS=16), indicating that the \textit{RPP} captures expert-activation patterns that generalize across tasks.

\subsubsection{Memory Cost}
\label{sec:memory_cost_analysis}
\begin{figure}
    \centering
    \includegraphics[width=0.9\linewidth]{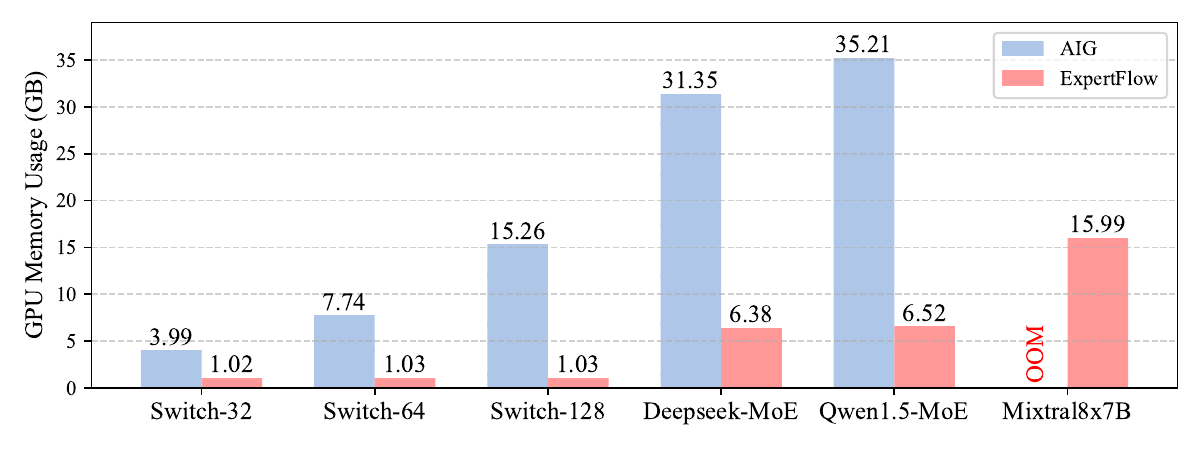}
    \caption{GPU memory usage (GB) of All-in-GPU (AIG) and our offloading-based system for different MoE models.}
    \label{fig:memory_result}
\end{figure}

We compare peak GPU memory against an All-In-GPU (\textbf{AIG}) baseline that retains all parameters in GPU (Fig.~\ref{fig:memory_result}). Our approach reduces memory by up to 93\% across Switch models (e.g., Switch-128: 15.26\,GB $\rightarrow$ 1.03\,GB), from 31.35\,GB to 6.38\,GB on Deepseek-MoE, and from 35.21\,GB to 6.52\,GB on Qwen1.5. Notably, Mixtral-8$\times$7B triggers OOM under AIG but completes with our system using only 15.99\,GB.

\subsection{Predictor Evaluation}\label{sec:predictor_eval}

\subsubsection{Routing Path Datasets} For each combination of task (i.e., Alpaca, XSUM, and WMT16) and MoE model, we construct a Routing Path Dataset (\textit{RPD}) to train and evaluate the routing predictor. We first sample 10{,}000 input sequences and run each sequence through the MoE model three times to collect diverse output tokens and the corresponding routing paths of both inputs and outputs. This yields 30{,}000 input–output–path triples per \textit{RPD}. Each \textit{RPD} is then split into training and test sets.

\subsubsection{Predictor Settings} To balance between performance and size, we conducted grid search on predictor architecture settings. The final \textit{RPP} has a feed-forward dimension of 2048, and a hidden size of 32, resulting in a 7.21 MB model size.

\subsubsection{Evaluation Metrics} 
To assess prediction performance in an MoE model with $L$ layers and $E$ experts per layer, we define the batch-level accuracy ($\mathbb{B}_{\text{acc}}$) as:

\begin{align*}
\mathbb{B}_{{\text{acc}}} &= \frac{1}{L} \sum_{i=0}^{L-1} \frac{\sum_{j=0}^{E-1} \mathbb{I}(R_{\text{batch}}[i, j] = 1\, \& \,\hat{R}_{\text{batch}}[i, j] = 1)}{\sum_{j=0}^{E-1} \mathbb{I}(R_{\text{batch}}[i, j] = 1)},
\end{align*}

\noindent where $R_{\text{batch}}$ and $\hat{R}_{\text{batch}}$ are the ground-truth and predicted batch-level routing matrices, respectively, both in $\{0,1\}^{L \times E}$. We apply a bitwise OR over the token-level routing paths $r_i \in \{0,1\}^{L \times E}$ to get the batch-level results $R_{\text{batch}} = r_1 \lor r_2 \lor \dots \lor r_n$.

\begin{figure}[!htb]
    \centering
    \subfigure[Switch32]{
    \includegraphics[width=.43\linewidth]{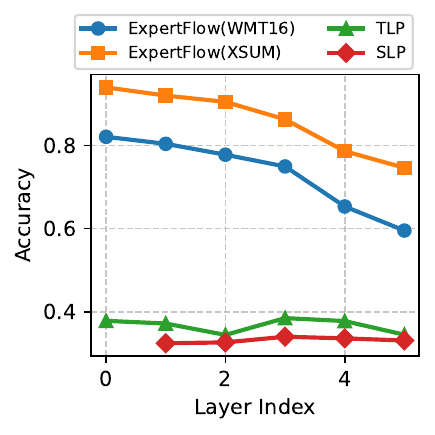}
    }
    \subfigure[Switch64]{
    \includegraphics[width=.43\linewidth]{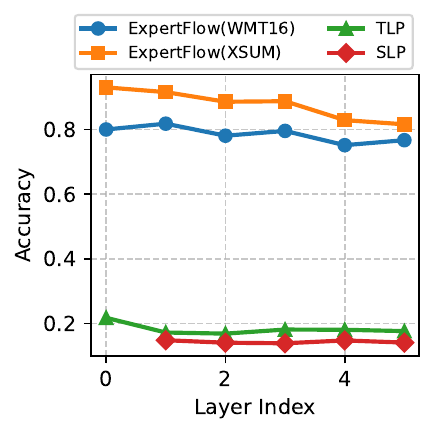}
    }
    \subfigure[Switch128]{
    \includegraphics[width=.43\linewidth]{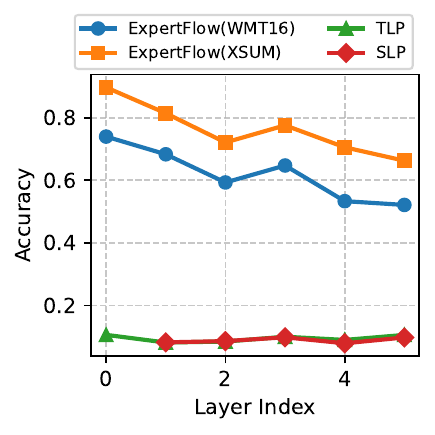}
    }
    \subfigure[Mixtral-8]{
    \includegraphics[width=.43\linewidth]{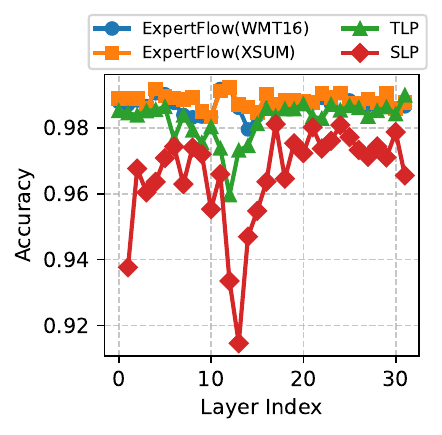}
    }
    \subfigure[Qwen1.5]{\includegraphics[width=.43\linewidth]{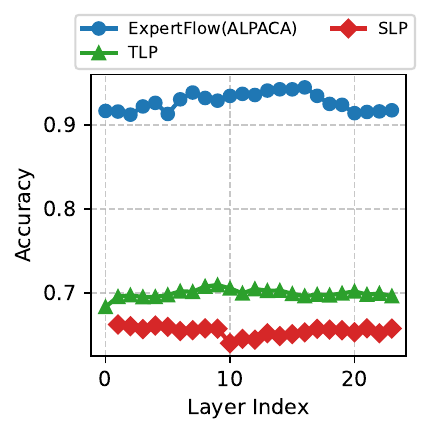}
    }
    \subfigure[Qwen1.5(Alpaca)]{
    \includegraphics[width=.43\linewidth]{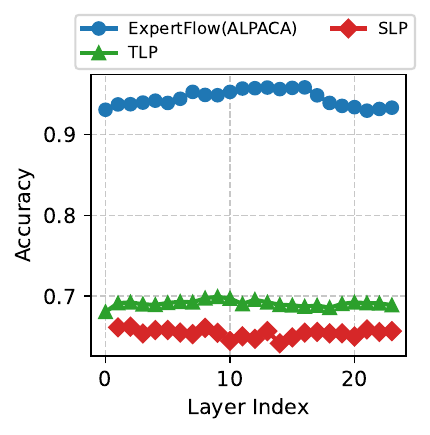}
    }
    \caption{Layer-wise expert prediction accuracy across MoE models. ``ExpertFlow(dataset)'' indicates our predictor trained on the specified dataset (e.g., XSUM or WMT16). \textcolor{orange}{Orange} and \textcolor{blue}{blue} curves represent in-domain and cross-domain prediction accuracy, respectively. Subfigures (a–e) report accuracy on XSUM; (f) shows in-domain results on Alpaca.  Our method consistently outperforms temporal-locality prediction (TLP) and spatial-locality prediction (SLP) baselines across all cases.}
    \label{fig:predictor_perf}
\end{figure}

\subsubsection{Predictor Performance} 

Fig.~\ref{fig:predictor_perf} presents the performance of our routing path predictor (\textit{RPP}) and two baselines: temporal-locality prediction (TLP), which uses the previous decoding step’s routing, and spatial-locality prediction (SLP), which relies on the previous layer’s expert assignment. Both are simple heuristics that lack context-awareness.

\textit{RPP} consistently achieves high prediction accuracy across MoE models and datasets. It outperforms both TLP and SLP in all settings, with over 90\% accuracy in most in-domain cases and only a modest drop (typically 5–10\%) in cross-domain scenarios. For example, on Switch64 (Fig.~\ref{fig:predictor_perf}b), our predictor maintains 80–90\% accuracy in both domains, while baselines remain below 20\%. This highlights \textit{RPP}’s strong generalization ability. Additionally, TLP consistently surpasses SLP, suggesting that temporal cues are more predictive than layer-wise locality.

Despite differing routing mechanisms, \textit{RPP} adapts well across models. On Switch models (a–c), accuracy declines with depth and expert count, reflecting increased uncertainty. Mixtral-8 (d) shows larger prediction variance for baselines due to its Top-2 routing, while \textit{RPP} remains above 90\% and more stable. On Qwen1.5 (e–f), our predictor achieves the highest accuracy (over 95\%), with minimal domain shift impact. These results demonstrate \textit{RPP}’s ability to capture complex and model-specific expert behaviors.

\subsection{Ablation Study}\label{sec:ablation_study}

\subsubsection{Expert Cache Hit Ratio.}

\begin{figure}[!htb]
    \centering
    \includegraphics[width=0.9\linewidth]{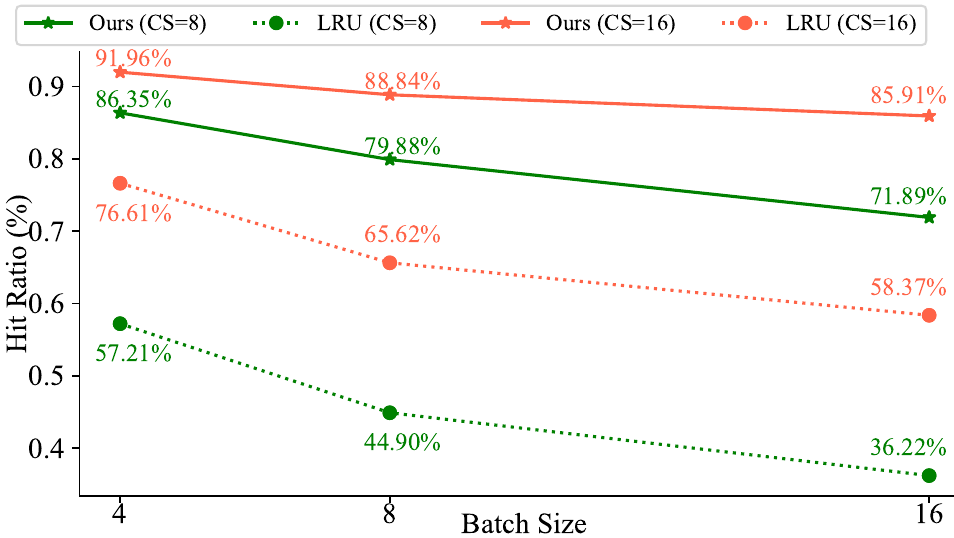}
    \caption{Comparison of expert cache hit ratio on Switch-32 under different batch size and cache size (CS). }
    \label{fig:hit-ratio}
\end{figure}

Fig.~\ref{fig:hit-ratio} compares expert cache hit ratios between LRU and our PLEC strategy on Switch-32. Our approach consistently outperforms LRU, with hit rates 15-36\% higher across all configurations. At CS=16, PLEC maintains high hit ratios (91.90\% to 85.91\%, only 6.05\% decrease) as batch size increases from 4 to 16, while LRU drops significantly (76.61\% to 58.37\%, an 18.24\% decline). The performance gap widens at larger batch sizes, with PLEC achieving 71.89\% hit ratio at CS=8/BS=16 compared to LRU's 36.22\%. This stability demonstrates our predictive approach's effectiveness in allocating cache resources based on anticipated expert patterns, particularly beneficial in high-throughput scenarios.

\subsubsection{Impact of Token Scheduler (TS) under Varying Expert Counts.} We evaluate the \textit{TS} on the Switch-series models to study its impact under different expert counts. This setup allows us to isolate the effect of \textit{TS} as model sparsity increases. As shown in Table~\ref{tab:ts_eval}, \textit{TS} yields consistent throughput improvements: $1.03\times$ on Switch-32, $1.15\times$ on Switch-64, and $1.17\times$ on Switch-128. While the absolute gains vary, a general trend of improved benefit with more experts is observed. This is attributed to the fact that, with a larger number of experts, token-to-expert assignments tend to become more fragmented. \textit{TS} alleviates this by rebatching tokens with similar routing paths, improving load balance and computational efficiency across experts.

\begin{table}[]
    \centering
    \caption{Impact of the \textit{Token Scheduler (TS)} on throughput.}
    \label{tab:ts_eval}
    \scalebox{0.9}{
    \begin{tabular}{c|c|c}\hline
    \multirow{2}{*}{Model} & \multicolumn{2}{c}{Throughput (tokens/second)}    \\\cline{2-3}
    & w/o TS & with TS \\\hline
    Switch-32 & 854.19   & 881.91 (1.03$\times$)    \\
    Switch-64 & 680.72   & 788.30 (1.15$\times$) \\
    Switch-128 & 628.36  & 735.35 (1.17$\times$)  \\
    \hline
    \end{tabular}
    }
\end{table}

\section{Conclusion}

We introduced {\name}, a unified system for memory-efficient MoE inference under tight GPU constraints. By integrating a \textbf{Routing Path Predictor}, a routing-aware \textbf{Token Scheduler}, and a predictive \textbf{Expert Cache Engine},  {\name} enables early expert planning, higher expert utilization, and reduced CPU–GPU transfers, yielding up to 93.72\% peak memory reduction and up to 10$\times$ throughput improvement across diverse MoE architectures when compared with strong offloading baselines. Accurate routing prediction also shows broader value for distributed expert placement, routing-guided pruning, and hierarchical caching, positioning {\name} as a foundation for future system-level co-design that incorporates predictive routing into scalable sparse model deployment and training.

\begin{acks}
\sloppy
This research is supported by the Career Development Fund (CDF) of the Agency for Science, Technology and Research (A*STAR), Singapore (No. C243512012); partially by the National Natural Science Foundation of China (NSFC) (No. 62302123) and the Shenzhen Science and Technology Program (Nos. KJZD20240903104103005, KJZD20230923114213027, KJZD20230923115113026); and partially by the National Research Foundation (NRF), Singapore, through the AI Singapore Programme under the project titled ``AI-based Urban Cooling Technology Development''(Award No. AISG3-TC-2024-014-SGKR).

\end{acks}
\newpage
\bibliographystyle{ACM-Reference-Format}
\bibliography{sample-base}

\end{document}